\documentclass[letterpaper]{article}
\usepackage{aaai2026}  
\usepackage{times}
\usepackage{helvet}
\usepackage{courier}
\usepackage[hyphens]{url}
\usepackage{graphicx}
\usepackage[round,authoryear]{natbib}
\usepackage{caption}
\usepackage{amsmath}    
\usepackage{tikz}         
\usetikzlibrary{arrows.meta,calc,shapes.geometric,positioning}  
\usepackage{algorithm}       
\usepackage{algorithmicx}    
\usepackage{algpseudocode}   
\usepackage{multirow}
\usepackage{verbatim}

\title{DIM: Enforcing Domain-Informed Monotonicity in Deep Neural Networks}
\author{Joshua Salim\textsuperscript{\rm 1}, Jordan Yu\textsuperscript{\rm 1}, Xilei Zhao\textsuperscript{\rm 1}\\
\textsuperscript{\rm 1}Department of Civil and Coastal Engineering, University of Florida\\
\{joshua.salim, jordanyu, xilei.zhao\}@ufl.edu}

\begin{document}

\maketitle

\begin{abstract}
While deep learning models excel at predictive tasks, they often overfit due to their complex structure and large number of parameters, causing them to memorize training data, including noise, rather than learn patterns that generalize to new data. To tackle this challenge, this paper proposes a new regularization method, i.e., \textit{Enforcing Domain-Informed Monotonicity in Deep Neural Networks} (DIM), which maintains domain-informed monotonic relationships in complex deep learning models to further improve predictions. Specifically, our method enforces monotonicity by penalizing violations relative to a linear baseline, effectively encouraging the model to follow expected trends while preserving its predictive power. We formalize this approach through a comprehensive mathematical framework that establishes a linear reference, measures deviations from monotonic behavior, and integrates these measurements into the training objective. We test and validate the proposed methodology using a real-world ridesourcing dataset from Chicago and a synthetically created dataset. Experiments across various neural network architectures show that even modest monotonicity constraints consistently enhance model performance. DIM enhances the predictive performance of deep neural networks by applying domain-informed monotonicity constraints to regularize model behavior and mitigate overfitting. 
\end{abstract}

\section{Introduction}
Deep learning has driven breakthroughs in many fields by extracting complex patterns from large datasets. However, their high capacity often leads to overfitting where networks memorize noise rather than learn generalizable trends. Domain experts rely on well-understood monotonic relationships (where increasing inputs never decrease outputs) to trust model predictions across healthcare, finance, energy, and transportation applications \cite{liu2020certified,lang2005monotonic,li2023unsupervised,nguyen2018energy,runje2023constrained}. For example, hospital readmission models should show non-decreasing risk with patient age \cite{li2023unsupervised}, credit scoring systems should reflect higher default probability with increased debt levels, and transportation demand models should show rising trip volumes with population density \cite{zhang2022ridesourcing,mokhtarian2001understanding}.

While deep neural networks often achieve high predictive accuracy, they often fail to respect simple monotonic relationships, such as ensuring that an increase in an input does not lead to a decrease in the output. In many practical settings, domain experts understand that certain input features should consistently influence predictions in a specific direction. However, conventional training methods do not inherently enforce these expectations, which can result in unpredictable or counterintuitive model behavior.

Current monotonicity research faces fundamental limitations in how violations are detected and penalized. Specialized architectures offer strong guarantees but sacrifice scalability and flexibility by fixing weight signs or employing monotonic building blocks \cite{archer1993learning,daniels2010monotone,you2017deep,runje2023constrained}. Early work by \cite{archer1993learning} introduced weight sign constraints for single-layer networks, while \cite{daniels2010monotone} extended this to deeper architectures, but these approaches complicate pipelines and reduce adaptability. More recent lattice-based methods \cite{you2017deep} and constrained ReLU architectures \cite{runje2023constrained} provide stronger theoretical guarantees but require specialized implementations that limit practical deployment. More critically, existing soft penalties using gradients or pairwise comparisons \cite{sill1996hints,gupta2019monotonicity} fundamentally lack an objective reference for what constitutes a violation. \cite{sill1996hints} pioneered monotonicity hints through Bayesian priors, while \cite{gupta2019monotonicity} proposed gradient-based regularization, but these methods penalize negative gradients or reversed pairs without establishing what the correct monotonic behavior should be. Domain-specific approaches in transportation modeling \cite{feng2024deep,kim2024flexible} have shown promise but remain tailored to discrete choice frameworks rather than general neural architectures. This absence of a violation baseline leads to arbitrary penalty scales and inconsistent constraint enforcement across different problem domains.

We propose \textit{\textbf{D}omain \textbf{I}nformed \textbf{M}onotonicity (DIM)}, a model-agnostic monotonicity penalty that fundamentally differs from prior work by establishing an explicit reference trend before measuring violations. \cite{shen2021modelagnosticgraphregularizationfewshot} Building on classical isotonic regression principles \cite{barlow1972isotonic} but adapted for modern deep learning, our approach addresses the core limitation of existing regularization methods. For each monotonic feature, we first fit a linear baseline to the model's current predictions, creating an interpretable benchmark representing the expected monotonic relationship. We then systematically compare the actual prediction changes against this baseline to quantify violation severity. This approach provides three key advantages over existing methods: (1) \textbf{Objective violation measurement} through comparison against a fitted trend rather than arbitrary thresholds, (2) \textbf{Interpretable penalty scaling} where violation magnitude directly corresponds to deviation from expected behavior, and (3) \textbf{Consistent enforcement} across different input regions and model architectures. Unlike pairwise methods \cite{sill1996hints} that only check local relationships or gradient methods \cite{gupta2019monotonicity} that lack global context, our linear reference provides a clear, interpretable standard for monotonic behavior while requiring no architectural modifications.

Our main contributions include:
\begin{itemize}
  \item We propose a monotonicity regularization framework that incorporates a linear trend reference, providing a structured basis for guiding model behavior and reducing overfitting.
  \item We design a differentiable penalty mechanism that selectively enforces monotonicity over designated features without modifying network architectures.
  \item We demonstrate the capability of monotonicity enforcement to improve model accuracy, revealing how domain-informed constraints improve generalization in noisy and low-signal settings.
  \item We conduct comprehensive experiments across synthetic and real-world travel demand datasets to uncover new insights into when and how monotonic constraints improve prediction accuracy and generalization.
\end{itemize}

\section{Related Work}
\subsection{Monotonicity in Statistics and Machine Learning}
Monotonicity constraints date back to classical statistical methods such as isotonic regression and order‐restricted maximum likelihood, which guarantee non‐decreasing fits when prior knowledge dictates such behavior \cite{barlow1972isotonic}. While these methods excel in low‐dimensional settings, their predictive power decreases with increasing feature complexity and nonlinearity. Recent work in transportation modeling has leveraged gravity and regression frameworks that assume monotonic trip‐generation relationships with variables like population and employment \cite{ortuzar2011modelling}, but these pre‐deep‐learning paradigms cannot capture high‐order interactions or latent spatial dependencies.

\subsection{Architectural Enforcement of Monotonicity}
To ensure that neural networks exhibit guaranteed monotonic behavior, researchers have developed specialized architectures that impose structural constraints.\cite{sill1998monotonic,sivaraman2020counterexample} One of the earliest approaches involves restricting network weights to be non‐negative in single‐hidden‐layer perceptrons, thereby ensuring the learned function remains monotonic across the entire input space \cite{archer1993learning}. Building on this, \cite{lang2005monotonic} introduced Monotonic Multi‐Layer Perceptrons (MON‐MLPs), which extend the same principle to deeper networks while maintaining their universal approximation capabilities. \cite{daniels2010monotone} further advanced this line of research by allowing partial monotonicity, constraining only specific input features while leaving others unrestricted. Another important development is the Deep Lattice Network, which combines piecewise‐linear calibrators and learned look‐up tables to enforce monotonicity across multiple inputs \cite{you2017deep}. While these architectures offer strong guarantees, they frequently incur exponential parameter growth or require custom layer implementations, limiting scalability and ease of integration. More recently, \cite{runje2023constrained} introduced ReLU‐based constrained architectures that encode monotonicity into activation patterns, but these too demand specialized design and can complicate standard machine learning workflows.

\subsection{Regularization and Model‐Agnostic Methods}
An alternative to architectural constraints is to add monotonicity penalties directly into the loss function. \cite{monteiro2022monotonic_reg} \cite{sill1996hints} pioneered Bayesian‐inspired “monotonicity hints” that discourage output reversals between pairs of inputs differing in one feature, and \cite{gupta2019monotonicity} proposed gradient‐based pairwise penalties to discourage negative partial derivatives. Although architecture‐agnostic, both methods enforce only local slope constraints and lack a global reference, often requiring careful hyperparameter tuning to avoid under or over regularization. \cite{liu2020certified} introduced Certified Monotonic Neural Networks, using mixed‐integer linear programming to verify monotonicity post‐training, but the MILP approach is computationally prohibitive in high dimensions. More recently, \cite{feng2024deep} and \cite{kim2024flexible} applied gradient regularization within discrete choice and travel‐demand models to improve behavioral realism; however, their techniques are tailored to specific shallow architectures and have not been demonstrated at scale in deep learning pipelines.

\section{Methodology}
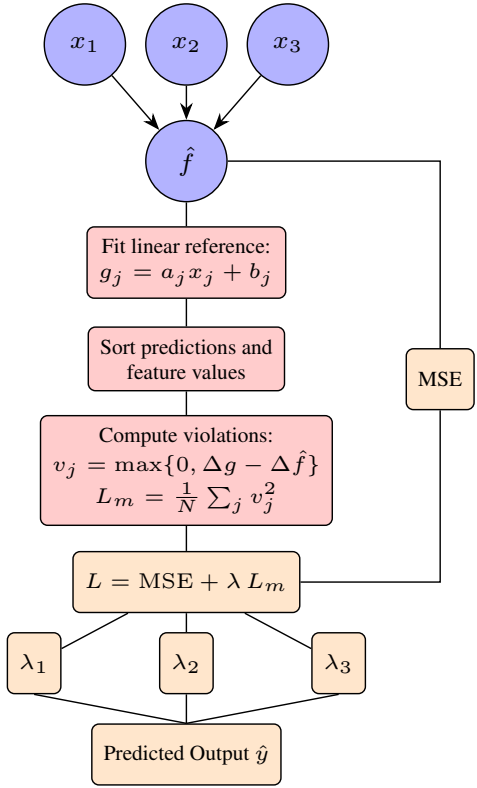
\begin{figure}[t]
  \centering
  \resizebox{0.35\textwidth}{!}{%
    \begin{tikzpicture}[
        >=Stealth,
        neuron/.style={
          circle, draw, minimum size=8mm, fill=blue!30, font=\scriptsize
        },
        box/.style={
          rectangle, draw, rounded corners=2pt, align=center,
          minimum height=0.6cm, fill=orange!20, font=\tiny
        },
        redbox/.style={
          rectangle, draw, rounded corners=2pt, align=center,
          minimum height=0.6cm, fill=red!20, font=\tiny
        }
      ]

      \node[neuron] (x1) at (-1,2.5) {$x_1$};
      \node[neuron] (x2) at ( 0,2.5) {$x_2$};
      \node[neuron] (x3) at ( 1,2.5) {$x_3$};

      \node[neuron] (f) at (0,1.35) {$\hat f$};
      \draw[->] (x1) -- (f);
      \draw[->] (x2) -- (f);
      \draw[->] (x3) -- (f);

      \node[redbox] (baseline) at (0,0.35) {%
        Fit linear reference:\\ $g_j = a_j x_j + b_j$
      };
      \node[redbox] (sort) at (0,-0.6) {%
        Sort predictions and\\ feature values
      };
      \node[redbox] (lm) at (0,-1.7) {%
        Compute violations:\\
        $v_j = \max\{0,\Delta g - \Delta \hat f\}$\\
        $L_m = \frac1N\sum_j v_j^2$
      };

      \draw[-] (f)         -- (baseline);
      \draw[-] (baseline) -- (sort);
      \draw[-] (sort)     -- (lm);

      \node[box] (mse) at (2.5,-0.8) {MSE};
      \draw[-] (f.east) -| (mse.north);

      \node[box] (loss) at (0,-2.8) {$L = \mathrm{MSE} + \lambda\,L_m$};
      \draw[-] (lm)        -- (loss);
      \draw[-] (mse.south) |- (loss.east);

      \node[box] (lam1) at (-1.5,-3.6) {$\lambda_1$};
      \node[box] (lam2) at ( 0,-3.6) {$\lambda_2$};
      \node[box] (lam3) at ( 1.5,-3.6) {$\lambda_3$};
      \draw[-] (loss) -- (lam1);
      \draw[-] (loss) -- (lam2);
      \draw[-] (loss) -- (lam3);

      \node[box] (output) at (0,-4.5) {Predicted Output $\hat{y}$};
      \draw[-] (lam1.south) -- (output.north);
      \draw[-] (lam2.south) -- (output.north);
      \draw[-] (lam3.south) -- (output.north);

    \end{tikzpicture}%
  }
  \caption{Overview of the DIM framework: Fit a linear baseline, sort predictions \& features, compute monotonicity violations, combine with MSE, and tune $\lambda$.}
  \label{fig:dim_flowchart_method}
\end{figure}

This section introduces the methodology for developing the monotonicity constraints and integrating them into the deep neural networks, helping to bridge the gap between predictive accuracy and domain knowledge without losing significant trade-offs from both sides. The overview of the proposed methodology is illustrated in Figure \ref{fig:dim_flowchart_method}.

\subsection{Notation and Problem Setup}

Let $X \in {R}^{N \times d}$ denote the input feature matrix for a batch of $N$ observations, where each observation has $d$ features. For a given observation $i$, we denote $X^{(i)} \in {R}^d$ as the feature vector and $X_j^{(i)}$ as the value of feature $j$ for observation $i$. Let $f: {R}^d \rightarrow {R}$ represent the true underlying function we wish to approximate, and $\hat{f}: {R}^d \rightarrow {R}$ denote our neural network model that produces predictions $\hat{f}^{(i)} = \hat{f}(X^{(i)})$ for each observation $i$.

We define $\mathcal{J}_M \subseteq \{1, 2, \ldots, d\}$ as the set of feature indices that are designated as monotonic based on domain knowledge. For each monotonic feature $j \in \mathcal{J}_M$, we expect that the relationship between feature $j$ and the output should be non-decreasing, i.e., if $X_j^{(i)} \leq X_j^{(k)}$, then ideally $\hat{f}^{(i)} \leq \hat{f}^{(k)}$ should hold.

\subsection{Establishing the Linear Baseline}
To quantify and enforce monotonic behavior, we first establish a linear regression baseline for each monotonic feature $j \in \mathcal{J}_M$.  This baseline serves as a reference to evaluate whether the model's predictions conform to an expected monotonic pattern \cite{magdonismail2008linear}. This process forms the top branch of our framework as depicted in Figure \ref{fig:dim_flowchart_method}.

For a given batch of $N$ observations, the optimal linear parameters are computed as follows:
\[
    a_j = \frac{\mathrm{Cov}(X_j, \hat{f})}{\mathrm{Var}(X_j)}, \quad b_j = \mu_{\hat{f}} - a_j \mu_{X_j},
\]
\noindent where the mean values of the input feature and the predicted output are given by:
\[
    \mu_{X_j} = \frac{1}{N}\sum_{i=1}^{N} X_j^{(i)}, \quad \mu_{\hat{f}} = \frac{1}{N}\sum_{i=1}^{N} \hat{f}^{(i)}.
\]

The covariance and variance terms are computed over the current batch:
\[
    \mathrm{Cov}(X_j, \hat{f}) = \frac{1}{N}\sum_{i=1}^{N} (X_j^{(i)} - \mu_{X_j})(\hat{f}^{(i)} - \mu_{\hat{f}}),
\]
\[
    \mathrm{Var}(X_j) = \frac{1}{N}\sum_{i=1}^{N} (X_j^{(i)} - \mu_{X_j})^2.
\]

The parameter $a_j$ represents the slope of the linear regression model, indicating the overall direction and magnitude of the relationship between feature $j$ and the predictions. For monotonic features, we expect $a_j > 0$ for monotonically increasing relationships. The intercept $b_j$ ensures that the linear approximation passes through the centroid $(\mu_{X_j}, \mu_{\hat{f}})$, preventing shifts that could bias the monotonicity assessment. By defining this linear baseline $g_j(x) = a_j x + b_j$ for each monotonic feature, we provide an objective measure of how a strictly monotonic function should behave, allowing for deviations to be identified and penalized.

\subsection{Sorting Predictions and Corresponding Feature Values}
To effectively compare the model's output with the established linear baseline, we first sort predictions in ascending order. This allows for a structured analysis of the model's behavior over increasing values of the monotonic feature. \cite{ng2004feature} The sorting process is represented by a permutation function $\sigma: \{1, 2, \ldots, N\} \rightarrow \{1, 2, \ldots, N\}$ such that:
\[
    \hat{f}^{(\sigma(1))} \leq \hat{f}^{(\sigma(2))} \leq \dots \leq \hat{f}^{(\sigma(N))}.
\]

After obtaining the permutation $\sigma$, we reorder both predictions and feature values to maintain consistency. We use the following notation for the sorted sequences:
\[
    \tilde{f}^{(i)} = \hat{f}^{(\sigma(i))}, \quad \tilde{x}_j^{(i)} = X_j^{(\sigma(i))}, \quad \forall i \in \{1, \dots, N\}.
\]

This sorting operation ensures that adjacent values in the sequence correspond to natural incremental changes in the prediction values. By structuring the data in this manner, we can systematically evaluate whether the prediction changes align with expected monotonic behavior relative to the changes in feature values.

\subsection{Linear Reference Predictions}
Once the predictions and feature values are sorted, we compute the linear reference predictions based on the established baseline model:
\vspace{-0.3em}
\[
    g(x)^{(i)} = a_j x_j^{(i)} + b_j, \quad \forall i \in \{1, \dots, N\}, \quad \forall j \in \mathcal{J}_M.
\]

These values represent the expected monotonic output for each observation, assuming a linear relationship. The linear reference predictions serve as a benchmark for evaluating how closely the model's predictions follow a monotonic trend.

\subsection{Measuring Deviations}
To quantify deviations from monotonicity, we calculate incremental changes between consecutive model predictions and compare them against the corresponding increments in the linear reference predictions:
\vspace{-0.2em}
\[
\begin{aligned}
  \Delta \hat{f}^{(i)} &= \hat{f}^{(i+1)} - \hat{f}^{(i)}, \\
  \Delta g(x)^{(i)} &= g(x)^{(i+1)} - g(x)^{(i)},
\end{aligned}
\]
\vspace{-0.7em}
\[ \quad \forall i \in \{1,\dots,N-1\}. \]

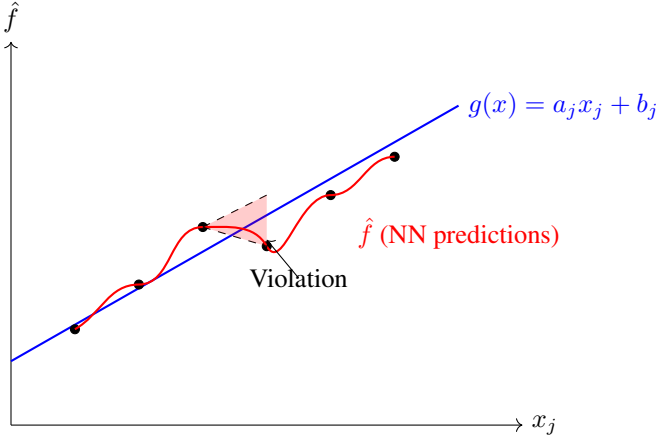
\begin{figure}[t]
    \centering
    \begin{tikzpicture}[scale=0.85]
        \draw[->] (0,0) -- (8,0) node[right] {$x_j$};
        \draw[->] (0,0) -- (0,6) node[above] {$\hat{f}$};
        
        \draw[thick, blue] (0,1) -- (7,5) node[right] {$g(x) = a_j x_j + b_j$};
        
        \filldraw[black] (1,1.5) circle (2pt);
        \filldraw[black] (2,2.2) circle (2pt);
        \filldraw[black] (3,3.1) circle (2pt);
        \filldraw[black] (4,2.8) circle (2pt); 
        \filldraw[black] (5,3.6) circle (2pt);
        \filldraw[black] (6,4.2) circle (2pt);
        
        \draw[thick, red] (1,1.5) to[out=30,in=180] (2,2.2) 
                         to[out=0,in=180] (3,3.1) 
                         to[out=0,in=120] (4,2.8) 
                         to[out=-60,in=180] (5,3.6)
                         to[out=0,in=180] (6,4.2);
        
        \draw[dashed, black] (3,3.1) -- (4,2.8);
        \draw[dashed, black] (3,3.1) -- (4,3.6); 
        \fill[red, opacity=0.2] (3,3.1) -- (4,2.8) -- (4,3.6) -- cycle;
        
        \node[red] at (7,3) {$\hat{f}$ (NN predictions)};
        \node at (4.5,2.3) {Violation};
        \draw[->] (4.5,2.3) -- (4,2.9);
    \end{tikzpicture}
    \caption{Visualization of monotonicity violation: The red curve represents neural network predictions, while the blue line shows the linear baseline. The shaded area highlights a region where the model violates the expected monotonic trend, as the prediction decreases despite an increase in feature value.}
    \label{fig:monotonicity_violation}
\end{figure}

If the change in the model's predictions is smaller than the change in the reference values, the model exhibits a violation of monotonicity. We formally define the violation function as:
\vspace{-0.3em}
\[ v_j^{(i)} = \max\{0, \Delta g(x)^{(i)} - \Delta \hat{f}^{(i)}\}. \]
This function captures instances where the neural network's predictions are less monotonic than the linear approximation, highlighting points where the model does not adhere to expected monotonic increases or decreases (see Figure \ref{fig:monotonicity_violation} for illustration).


\subsection{Monotonicity Penalty Calculation}
To enforce monotonicity, we define a penalty function that sums squared deviations across all observations for each monotonic feature:

\[
    P_j = \sum_{i=1}^{N-1}(v_j^{(i)})^2.
\]

Squaring the violations ensures that larger deviations receive higher penalties, thereby prioritizing the correction of significant monotonicity violations during training.

\subsection{Total Monotonicity Penalty}
To normalize the penalty across different batch sizes, we compute the overall monotonicity penalty as:
\vspace{-0.4em}
\[
    \mathcal{L}_{\text{m}}(\hat{f}, X) = \frac{1}{N}\sum_{j \in \mathcal{J}_M}P_j.
\]
\vspace{-0.3em}
This normalization ensures that the scale of the penalty remains consistent, regardless of the number of training samples.

\subsection{Final Objective Function}
\vspace{-0.3em}
The monotonicity penalty is added into the model's training loss function to balance predictive accuracy with monotonicity enforcement:
\vspace{-0.4em}
\[
    \mathcal{L}(\hat{f}, y) = \text{MSE} + \lambda \cdot \mathcal{L}_{\text{m}}(\hat{f}, X).
\]
\vspace{-0.2em}
\noindent where the hyperparameter $\lambda$ determines the weight assigned to monotonicity enforcement. A larger $\lambda$ strengthens monotonicity constraints, ensuring stricter adherence to expected relationships, while a smaller $\lambda$ allows more flexibility, prioritizing accuracy. The proper tuning of this hyperparameter is essential to achieving the optimal balance between prediction quality while still remaining adherence to domain knowledge. The full procedure for computing the penalty is summarized in Algorithm \ref{alg:monotonicity_penalty}.
\vspace{-0.2em}

\noindent
\begin{minipage}{\linewidth}
\begin{algorithm}[H]
\caption{Calculating Monotonicity Penalty}
\label{alg:monotonicity_penalty}
\begin{algorithmic}[1]
\Require Neural network predictions $\hat{f}$, Features $X$, Set of monotonic features $\mathcal{J}_M$
\Ensure Monotonicity penalty $\mathcal{L}_{\text{m}}(\hat{f}, X)$
\State $\mathcal{L}_{\text{m}} \gets 0$
\For{each $j \in \mathcal{J}_M$}
    \State Compute $a_j \gets \frac{\mathrm{Cov}(X_j, \hat{f})}{\mathrm{Var}(X_j)}$
    \State Compute $b_j \gets \mu_{\hat{f}} - a_j \mu_{X_j}$
    \State Sort predictions: $\hat{f}^{(1)} \leq \hat{f}^{(2)} \leq \dots \leq \hat{f}^{(N)}$
    \State Reorder feature values accordingly: $x_j^{(i)} = X_j^{(\sigma(i))}$
    \State $P_j \gets 0$
    \For{$i \gets 1$ to $N-1$}
        \State $\Delta \hat{f}^{(i)} \gets \hat{f}^{(i+1)} - \hat{f}^{(i)}$
        \State $\Delta g(x)^{(i)} \gets a_j(x_j^{(i+1)} - x_j^{(i)})$
        \State $v_j^{(i)} \gets \max\{0, \Delta g(x)^{(i)} - \Delta \hat{f}^{(i)}\}$
        \State $P_j \gets P_j + (v_j^{(i)})^2$
    \EndFor
    \State $\mathcal{L}_{\text{m}} \gets \mathcal{L}_{\text{m}} + P_j$
\EndFor
\State \Return $\frac{\mathcal{L}_{\text{m}}}{N}$
\end{algorithmic}
\end{algorithm}
\end{minipage}

\renewcommand{\arraystretch}{1.0}

\section{Experiments}

\subsection{Experimental Setup}
We validate our monotonicity‐constrained training on two different datasets, including a synthetic and a real-world dataset. To evaluate performance under controlled conditions, we generate a synthetic benchmark of 5,000 samples. For each sample, five independent features are drawn uniformly and then passed through simple increasing transformations. We inject piecewise constant “bumps” into each feature to introduce realistic, localized deviations before adding Gaussian noise to form the final target. By design, every synthetic feature maintains a perfect monotonic relationship with the output, allowing us to confirm that our penalty operates correctly on any tabular data with user‐defined monotonic constraints.

The second is a real-world ridesourcing dataset from Chicago, Illinois, comprising 67,498 origin–destination pairs recorded between November 1, 2018 and March 31, 2019. \cite{chi_tnp_trips_2018_2022} Each observation includes six trip-level statistics (the median and standard deviation of fare, distance, and duration), binary indicators for airport and downtown involvement, and an extensive suite of census-tract attributes covering socioeconomic, built-environment, transit, and employment measures. From this feature pool of our dataset, we identify five inputs: home-to-work commuters, work-to-home commuters, origin employment density, destination employment density, and the downtown-to-downtown flag. According to the domain knowledge and empirical findings, all these features should have a non-decreasing relationship with regard to trip volume \cite{zhang2024traveldemand}. These form our monotonic feature set, \(\mathcal{J}_M\). (See Appendix section A.1)

For both datasets, we apply an 80/20 train/test split and normalize all numerical inputs using min--max scaling computed separately on each split. We then train four deep neural network architectures, including (1) a single-layer ANN with 128 ReLU units, (2) an MLP-3 network with layers of 128, 64, and 32 units plus 20\% dropout, (3) an MLP-5 network with five hidden layers (256, 128, 64, 32, 16 units) and 20\% dropout, and (4) a three-layer Conv1D model (128$\rightarrow$64$\rightarrow$32 filters, kernel size 3, followed by global average pooling). These models were selected to represent a range of depths and architectural types commonly used in tabular data settings, allowing us to test how monotonicity constraints perform across different learning scenarios. All models train for up to 50 epochs under the Adam optimizer (learning rate $10^{-3}$), with a batch size of 256 and early stopping based on validation MSE.\cite{glorot2010understanding}

At each training step, we compute a monotonicity penalty as follows: for every feature in \(\mathcal{J}_M\), we fit a one-dimensional ordinary least squares line to the sorted batch predictions versus feature values, then measure any downward deviations from this linear trend. Squared deviations are averaged across features to produce the monotonicity regularization term.

\begin{table*}[t]
  \centering
  \footnotesize
  \setlength{\tabcolsep}{2pt}
  \renewcommand{\arraystretch}{0.95}
  \resizebox{0.85\textwidth}{!}{%
    \begin{tabular}{|l|l|c|c|c|c|c|}
      \hline
      \textbf{Feature} & \textbf{Model}
        & \textbf{Baseline MSE}
        & \textbf{Best MSE ($\lambda$)}
        & \textbf{\% Drop in MSE}
        & \textbf{\% Drop in MAE}
        & \textbf{\% Drop in MAPE} \\
      \hline
      \multirow{4}{*}{$x_1$}
        & ANN   & 0.32057 & 0.31106\,(\,0.2\,) &  2.97\% &  0.89\% & 14.02\% \\
        & CNN   & 0.24235 & 0.21205\,(\,0.6\,) & 12.50\% &  7.35\% & 28.75\% \\
        & MLP3  & 0.26765 & 0.23121\,(\,0.4\,) & 13.62\% &  7.04\% & 27.02\% \\
        & MLP5  & 0.28392 & 0.24167\,(\,0.2\,) & 14.88\% &  9.46\% & 59.38\% \\
      \hline
      \multirow{4}{*}{$x_2$}
        & ANN   & 0.32057 & 0.30870\,(\,0.8\,) &  3.70\% &  1.16\% &  6.37\% \\
        & CNN   & 0.24235 & 0.21381\,(\,0.2\,) & 11.78\% &  7.27\% & 23.50\% \\
        & MLP3  & 0.26765 & 0.21875\,(\,0.4\,) & 18.28\% &  9.48\% & 40.98\% \\
        & MLP5  & 0.28392 & 0.24653\,(\,1.0\,) & 13.18\% &  7.10\% & 62.93\% \\
      \hline
      \multirow{4}{*}{$x_3$}
        & ANN   & 0.32057 & 0.30964\,(\,1.0\,) &  3.41\% &  1.44\% & 10.21\% \\
        & CNN   & 0.24235 & 0.20890\,(\,0.6\,) & 13.80\% &  7.48\% & 35.27\% \\
        & MLP3  & 0.26765 & 0.21521\,(\,1.0\,) & 19.59\% & 10.46\% & 23.91\% \\
        & MLP5  & 0.28392 & 0.26161\,(\,1.0\,) &  7.86\% &  4.83\% & 56.36\% \\
      \hline
    \end{tabular}%
  }
  \caption{Best percentage drops in MSE, MAE, and MAPE across $\lambda>0$ for each feature and model.}
  \label{tab:mse_mae_mape_drops}
\end{table*}

\subsection*{Synthetic Dataset}
To evaluate our monotonicity‐constrained training in a controlled setting, we generate a synthetic dataset with \(N = 5{,}000\) observations and four independent features:
\[
x_1 \sim \mathcal{U}(0,200),\quad
x_2 \sim \mathcal{U}(0,50)\\[0.5em],
\]
\\ \vspace{-1.3em}
\[
x_3 \sim \mathcal{U}(0,150),\quad
x_4 \sim \mathcal{U}(0,100).
\]

Each feature is mapped through a strictly increasing baseline transform:
\[
f_1(x_1) = 0.5\,x_1,\quad
f_2(x_2) = 1.2\,\sqrt{x_2}\\[0.5em]
\]
\[
f_3(x_3) = 2.0\,\log\bigl(1 + x_3\bigr),\quad
f_4(x_4) = -0.8\,x_4.
\]
Note that \(x_4\) is included as a non‐monotonic feature to introduce additional variation.

To introduce localized departures from strict monotonicity, we partition each \(x_j\) into \(B\) equal‐width bins and add a random offset:
\begin{align*}
b &= \left\lfloor \frac{B\,(x_j - \min x_j)}{\max x_j - \min x_j}\right\rfloor,\\
\Delta_j(x_j) &= \delta_{j,b},\quad \delta_{j,b}\sim\mathcal{N}(0,\sigma_j^2).
\end{align*}
These “bumps” mimic realistic, piecewise variations while preserving the overall monotonic trend and test the robustness of our penalty against local perturbations.

Finally, we add Gaussian noise \(\varepsilon_i\sim\mathcal{N}(0,10^2)\) and assemble the target:
\begin{align*}
y_i &= f_1(x_{i,1}) + \Delta_1(x_{i,1})
      + f_2(x_{i,2}) + \Delta_2(x_{i,2})\\
    &\quad{}+ f_3(x_{i,3}) + \Delta_3(x_{i,3})
      + f_4(x_{i,4}) + \Delta_4(x_{i,4})
      + \varepsilon_i.
\end{align*}
The full dataset \(\{(x_{i,1},x_{i,2},x_{i,3},x_{i,4},y_i)\}_{i=1}^N\) is saved and verified to retain strong positive correlations between each \(x_j\) and \(y\), confirming the embedded monotonic structure (see Appendix A.3).

\subsection{Lambda Tuning}
Choosing the penalty weight \(\lambda\) requires balancing the competing objectives of predictive accuracy and monotonic compliance. We therefore employ a two-stage validation strategy on our 80/20 split. First, we perform a coarse grid search over \(\lambda \in \{0.0,0.2,0.4,0.6,0.8,1.0\}\), recording both validation MSE and a monotonicity-compliance score (the proportion of adjacent prediction pairs that respect the expected linear trend).\cite{bergstra2012random} In the second stage, we select the \(\lambda\) value that minimizes validation error without significantly sacrificing compliance. This procedure consistently identifies a moderate regularization strength that generalizes well across architectures and datasets, avoiding both underregularization (which fails to enforce monotonicity) and overregularization (which degrades predictive performance).

\begin{table*}[t]
  \centering
  \normalsize
  \setlength{\tabcolsep}{3.5pt}
  \renewcommand{\arraystretch}{0.90}
  \resizebox{\textwidth}{!}{%
    \begin{tabular}{|l|l|c|c|c|c|c|}
      \hline
      \textbf{Feature} & \textbf{Model}
        & \textbf{Baseline MSE}
        & \textbf{Best MSE ($\lambda$)}
        & \textbf{\% Drop in MSE}
        & \textbf{\% Drop in MAE}
        & \textbf{\% Drop in MAPE} \\
      \hline
      \multirow{4}{*}{Commuters\_HW}
        & ANN   & 14.42935 & 12.33274\,(\,0.2\,) & 14.53\% &  -1.01\% &  -1.26\% \\
        & CNN   & 38.83416 & 31.85276\,(\,0.6\,) & 17.97\% &  -2.98\% &  -9.10\% \\
        & MLP3  & 17.05428 & 15.08996\,(\,1.0\,) & 11.52\% &   3.86\% &   4.36\% \\
        & MLP5  & 23.73539 & 19.00458\,(\,0.4\,) & 19.94\% &  41.07\% &  58.48\% \\
      \hline
      \multirow{4}{*}{Commuters\_WH}
        & ANN   & 14.42935 & 10.76013\,(\,0.4\,) & 25.43\% &  11.96\% &  17.84\% \\
        & CNN   & 38.83416 & 29.32747\,(\,1.0\,) & 24.47\% &  -1.88\% & -12.98\% \\
        & MLP3  & 17.05428 & 17.43412\,(\,0.8\,) & -2.27\% &  -1.10\% &  -1.73\% \\
        & MLP5  & 23.73539 & 17.55623\,(\,0.8\,) & 26.05\% &  18.10\% &  20.83\% \\
      \hline
      \multirow{4}{*}{EmpDen\_Des}
        & ANN   & 14.42935 & 14.27516\,(\,0.2\,) &  1.07\% &   8.52\% &  15.17\% \\
        & CNN   & 38.83416 & 34.35831\,(\,0.8\,) & 11.53\% &  -3.99\% & -15.55\% \\
        & MLP3  & 17.05428 & 16.41066\,(\,0.2\,) &  3.77\% &   9.67\% &  21.52\% \\
        & MLP5  & 23.73539 & 15.15645\,(\,0.4\,) & 36.14\% &  36.97\% &  53.07\% \\
      \hline
      \multirow{4}{*}{EmpDen\_Ori}
        & ANN   & 14.42935 & 12.70460\,(\,0.4\,) & 11.95\% &   8.74\% &  12.52\% \\
        & CNN   & 38.83416 & 31.93617\,(\,1.0\,) & 17.76\% &  -3.02\% & -13.67\% \\
        & MLP3  & 17.05428 & 15.46502\,(\,0.6\,) &  9.32\% &   2.38\% &   4.88\% \\
        & MLP5  & 23.73539 & 16.40746\,(\,0.2\,) & 30.88\% &  36.07\% &  49.27\% \\
      \hline
      \multirow{4}{*}{downtown\_downtown}
        & ANN   & 14.42935 & 13.94258\,(\,0.8\,) &  3.37\% &   3.27\% &   3.49\% \\
        & CNN   & 38.83416 & 34.54675\,(\,0.8\,) & 11.04\% &  -7.52\% & -21.56\% \\
        & MLP3  & 17.05428 & 16.53880\,(\,1.0\,) &  3.02\% &  -0.06\% &   0.58\% \\
        & MLP5  & 23.73539 & 17.65814\,(\,0.2\,) & 25.61\% &  41.12\% &  57.53\% \\
      \hline
    \end{tabular}%
  }
  \caption{Best percentage drops in MSE, MAE, and MAPE across $\lambda>0$ for each feature and model on the Chicago dataset. (A negative \% drop means an increase in the metric.)}
  \label{tab:mse_mae_mape_drops_chicago}
\end{table*}

\renewcommand{\arraystretch}{1.0}

\renewcommand{\arraystretch}{1.0}  

\subsection{Results on the Synthetic Dataset}
Our initial evaluation on the controlled synthetic benchmark demonstrates that the monotonicity penalty works when the data‐generating process already obeys strict monotonicity. As shown in Table 1 (see Table 4 in Appendix for full results), baseline test‐set MSE values lie between 0.242 and 0.320 in the four features. Introducing the penalty yields reductions of up to 19.6\% are observed (e.g., MLP3 on \(x_3\) at \(\lambda=1.0\)), while simpler models such as ANN and CNN maintain nearly invariant error throughout \(\lambda\in[0.0,1.0]\). These results confirm that our regularization does not worsen the accuracy of the predictions when the model is already naturally monotonic.

\subsection{Results on the Chicago Dataset}
We then validated our approach on the more complex real‑world Chicago ridesourcing dataset. Table 2 (see Table 3 in Appendix for full results) summarizes the best MSE reductions obtained across the four architectures and five monotonic features. The most pronounced gain occurs for the density of origin employment with MLP 5, where the MSE drops from 24.18 to 15.89 at \(\lambda=0.8\) (–34.30\%). Other notable improvements include CNN on EmpDen\_Des (-27.87\% at \(\lambda=0.6\)), MLP5 on downtown-downtown pairs (-27. 30\% at \(\lambda=0.4\)) and MLP5 on Commuters\_WH (-26.05\% at \(\lambda=0.8\)). Even simple ANN benefits, with a reduction of 28.83\% on Commuters\_HW at \(\lambda=0.4\). In fact, every model–feature combination sees at least one \(\lambda>0\) that lowers the test MSE, demonstrating that modest monotonic constraints can substantially enhance predictive performance in noisy real-world settings.

\textbf{Inconsistent Results Across Metrics:} While MSE improvements are consistent across all model-feature combinations, we observe mixed results for MAE and MAPE metrics. For instance, CNN models show substantial MSE reductions (11-18\%) but often exhibit increased MAE and MAPE values, particularly for Commuters\_WH (-1.88\% MAE, -12.98\% MAPE) and EmpDen\_Des (-3.99\% MAE, -15.55\% MAPE). This inconsistency suggests that while our monotonicity constraints help reduce large prediction errors (reflected in MSE), they may not uniformly improve all aspects of prediction quality in complex, noisy real-world data. The discrepancy between metrics likely comes from the messy, real-world nature of the Chicago dataset, where different areas have varying noise levels and complex relationships that our simple linear baseline cannot fully capture.

\subsection{Why This Method Works}
Our consistent improvements across both the synthetic and real-world datasets result from four methodological design elements that directly support our main contributions.

\textbf{Linear Trend Reference Framework:} We anchor monotonicity penalties to a fitted linear baseline instead of using arbitrary pairwise or gradient-based constraints. This strategy provides a clear, interpretable benchmark that captures the expected directional trend while remaining efficient and mathematically sound. Unlike earlier methods lacking a clear violation baseline \cite{sharma2020testing}, our approach accurately measures and penalizes deviations, addressing our first contribution of guiding model behavior with a principled reference. \cite{liu2020certified}

\textbf{Architecture-Agnostic Integration:} We integrate the monotonicity penalty into the loss function, allowing it to work with existing training pipelines without specialized layers, custom architectures, or post-training modifications \cite{shen2021modelagnosticgraphregularizationfewshot}. This approach preserves original network designs, from simple ANNs to complex MLPs and CNNs, while enforcing monotonicity on expert-selected features, fulfilling our second contribution of combining black‑box models with accurate predictions without changing their structure.

\textbf{Regularization-Driven Accuracy Improvements:} Modest monotonicity constraints serve as effective regularizers in noisy settings with low signal. \cite{monteiro2022monotonic_reg} By enforcing domain-informed relationships, we reduce overfitting to spurious patterns and guide the model toward generalizable trends \cite{dash2022review}. This results in up to 30\% reductions in MSE, particularly in deeper MLPs and real-world data sets. This confirms our third contribution to the generalization benefits of monotonicity enforcement.

\textbf{Systematic Constraint Selection and Tuning:} We use a two-stage $\lambda$ selection process along with expert guidance to apply constraints judiciously. A principled grid search balances enforcement strength against predictive performance, showing that moderate penalty weights consistently yield the best results.\cite{bergstra2012random} This systematic method supports our fourth contribution of offering practical advice for applying monotonic constraints.

Together, these elements transform domain knowledge into actionable rules, enabling deep models to achieve both high predictive accuracy and adherence to expected monotonic trends. Our framework provides structure without sacrificing flexibility, making it a valuable tool for creating trustworthy AI systems in domains that rely on expert knowledge.

\subsection{Limitations}
While our proposed method for enforcing monotonicity in deep models offers a flexible and model‑agnostic approach with promising results, it is not without limitations.

\textit{First}, our penalty is based on a linear reference approximation, which, although interpretable, may not fully capture complex domain-specific monotonic relationships that exhibit nonlinear trends. In such cases, enforcing alignment with a linear baseline could introduce bias, particularly in regions of the feature space where the true relationship is convex, concave, or piecewise monotonic. \cite{richman2024icenet}

\textit{Second}, our experiments are limited to tabular data and feed‑forward or convolutional neural networks. While the formulation is, in principle, extensible to other modalities such as time series, sequences, or graphs, this extension is not explicitly validated in our current study. \cite{simonyan2013deep} The generalization of monotonicity enforcement across data structures and application domains remains an important direction for future research.

\textit{Third}, our method evaluates monotonicity violations over batch‑wise predictions, which may lead to instability or inconsistencies during training on small batches or in the presence of high variance in feature distributions. \cite{masters2018smallbatch} While we mitigate this with careful batch sizing and normalization, extending the method to account for global monotonicity across the entire dataset could improve robustness.

\textit{Fourth}, our approach enforces monotonicity only between adjacent prediction pairs after sorting, which may fail to detect non-local violations such as dips and rebounds across multiple steps. For example, in the sequence $\hat{f}^{(1)} < \hat{f}^{(2)} < \hat{f}^{(3)} > \hat{f}^{(4)} < \hat{f}^{(5)}$, each adjacent pair appears monotonic, but the full sequence is not. This local view can miss broader inconsistencies that undermine global monotonicity. Future work could extend our method by incorporating multi-step or cumulative violation detection, leveraging sliding windows, isotonic regression, or hierarchical penalties to capture both local and global trends more robustly.

\textit{Fifth}, while our method consistently improves MSE across models and features, results for MAE and MAPE are mixed, particularly on the real-world Chicago dataset. Some configurations even see worse MAE/MAPE despite improved MSE. This likely stems from the dataset’s random noise and nonlinear relationships, where reducing large errors (which dominate MSE) does not guarantee better metrics. These inconsistencies highlight the limitations of a fixed linear baseline in complex domains and point to the potential benefits of adaptive or nonlinear reference trends.

Despite these limitations, our results show that properly formulated monotonicity constraints can improve prediction accuracy in neural networks, particularly in domains where domain logic and trust are critical. The adjacent-pair approach offers an efficient and sound foundation that could support more advanced global monotonicity methods in future research.

\section{Conclusion}
To summarize, we propose DIM, a model-agnostic monotonicity penalty that establishes a linear baseline to measure and penalize violations of domain-informed constraints. It advances our understanding of monotonic neural networks in complex datasets, helps researchers with a powerful yet easy to adapt tool for trustworthy demand forecasting, and lays the groundwork for a new generation of interpretable, reliable, and socially responsive AI systems. Our approach provides clear violation constraints while requiring no architectural modifications. Experiments on synthetic and real-world Chicago ridesourcing data show consistent MSE reductions of 20-30\% across multiple neural architectures. This work bridges black-box deep neural networks' performance with domain knowledge, providing researchers a practical tool for building trustworthy, deep learning models applicable across transportation, healthcare, finance, and many more domains.

\bibliographystyle{aaai2026}
\bibliography{references}

\clearpage




\section*{Appendix}

\subsection*{A.1 Chicago Travel Dataset}
To demonstrate the effectiveness of our monotonicity-regularized training approach, we use a real-world ridesourcing dataset from the Chicago Data Portal. The dataset spans from November 1, 2018 to March 31, 2019, and consists of $67{,}498$ unique origin–destination (OD) pairs between census tracts in Chicago, IL. Each OD record includes an 11-digit origin and destination tract identifier, along with the total number of trips recorded between the two tracts over the study period.

Trip-level attributes are captured using six statistical descriptors: median fare and fare standard deviation, median distance in miles and its standard deviation, and median travel time in seconds along with its standard deviation. Two binary indicators denote whether either end of the trip involves an airport tract or lies within the city's designated downtown area.

In addition to trip-level metrics, the dataset includes rich contextual features for both the origin and destination tracts. These features include:
\begin{itemize}
    \item \textbf{Socioeconomic composition:} transit usage (\texttt{pcttransit}), renter occupancy (\texttt{pctrentocc}), prevalence of single-family housing (\texttt{pctsinfam2}), and car ownership rates (\texttt{carown}).
    \item \textbf{Income and demographics:} shares of low-, moderate-, and middle-income residents (\texttt{pctlowinc}, \texttt{pctmodinc}, \texttt{pctmidinc}); youth indicator (\texttt{young2}); gender (\texttt{pctmale}); and racial/ethnic composition (\texttt{pctwhite}, \texttt{pcthisp}, \texttt{pctasian}).
    \item \textbf{Population and built environment:} population density (\texttt{popden}), crime density (\texttt{CrimeDen}), point-of-interest density (\texttt{InterstDen}), road network density (\texttt{RdNetwkDen}), and Walk Score (\texttt{Walkscore}).
    \item \textbf{Transit accessibility:} service hours for bus and rail routes (\texttt{SerHourBusRoutes}, \texttt{SerHourRailRoutes}), percentages of tract area within buffer zones (\texttt{PctBusBuf}, \texttt{PctRailBuf}), and densities of bus stops and rail stations (\texttt{BusStopDen}, \texttt{RailStationDen}).
    \item \textbf{Employment and commuting:} overall and retail employment densities (\texttt{EmpDen}, \texttt{EmpRetailDen}), workforce composition by wage and education level (\texttt{PctWacWorker54}, \texttt{PctWacLowMidWage}, \texttt{PctWacBachelor}), and bidirectional commuter flow counts (\texttt{Commuters HW}, \texttt{Commuters WH}).
\end{itemize}
An additional flag identifies OD pairs in which both tracts fall within the downtown boundary.

All census-tract-level features are joined using the 2015–2019 American Community Survey (ACS) five-year estimates. Features are standardized to $[0,1]$ using min–max normalization applied independently to the training and test splits.

\subsection*{A.2 Monotonic Feature Selection and Processing}
\vspace{-0.3em}
A subset of features is designated as monotonic based on urban mobility theory and empirical evidence. These include variables such as \texttt{EmpDen}, \texttt{Commuters HW}, and \texttt{pcttransit}, which are expected to have a non-decreasing relationship with trip generation. This designation is essential for ensuring that model predictions remain interpretable and aligned with real-world behavior.

Missing records are removed prior to training to ensure data consistency. The final dataset is randomly partitioned into 80\% training and 20\% testing subsets, and normalization is performed separately on each split. These steps prepare the data for use with our monotonicity penalty, ensuring that the model can leverage structural constraints while maintaining predictive performance.

\subsection{A.4 Model Configurations}  
We compare four architectures trained with monotonicity penalty weights $\lambda\in\{0.0,0.2,0.4,0.6,0.8,1.0\}$:
\begin{itemize}
  \item \textbf{ANN}: single hidden layer (128 ReLU), no convolution.
  \item \textbf{MLP‐3}: three hidden layers [128,64,32] ReLU + dropout(0.2).
  \item \textbf{MLP‐5}: five hidden layers [1024,512,256,128,64] ReLU + dropout(0.2).
  \item \textbf{CNN}: three Conv1D layers (128,64,32 filters, kernel=3) + GlobalAveragePooling1D.
\end{itemize}
All models are trained for 500 epochs with Adam (learning rate $10^{-3}$), batch size 256, and early stopping on validation MSE.

\subsection{A.5 Evaluation Metrics}  
We measure predictive performance using Mean Squared Error (MSE), Mean Absolute Error (MAE), and Mean Absolute Percentage Error (MAPE):
\[
\mathrm{MSE} = \frac{1}{N}\sum_{i=1}^N (\hat y_i - y_i)^2,\quad
\mathrm{MAE} = \frac{1}{N}\sum_{i=1}^N |\hat y_i - y_i|,\quad
\] \\
\[
\mathrm{MAPE} = \frac{100}{N}\sum_{i=1}^N \frac{|\hat y_i - y_i|}{y_i}\,.
\]

\begin{table*}[!t]
  \centering
  \small
  \resizebox{\textwidth}{!}{%
    \begin{tabular}{|l|l|ccc|ccc|ccc|}
      \hline
      \textbf{Feature} & \textbf{Model}
        & \multicolumn{3}{c|}{$\lambda=0.0$}
        & \multicolumn{3}{c|}{$\lambda=0.2$}
        & \multicolumn{3}{c|}{$\lambda=0.4$} \\
      \cline{3-11}
      & & MSE & MAE & MAPE & MSE & MAE & MAPE & MSE & MAE & MAPE \\
      \hline
      \multirow{4}{*}{Commuters\_HW}
      & ANN   & 14.429 & 1.281 & 0.797 & 12.332 & 1.298 & 0.807 & 20.976 & 1.591 & 1.096 \\
      & CNN   & 38.834 & 2.025 & 1.137 & 37.835 & 2.067 & 1.220 & 35.913 & 2.074 & 1.222 \\
      & MLP3  & 17.054 & 1.272 & 0.828 & 19.779 & 1.534 & 1.069 & 17.815 & 1.144 & 0.618 \\
      & MLP5  & 23.735 & 1.902 & 1.464 & 26.229 & 1.614 & 1.102 & 19.004 & 1.121 & 0.608 \\
      \hline
      \multirow{4}{*}{Commuters\_WH}
      & ANN   & 14.429 & 1.281 & 0.797 & 20.364 & 1.690 & 1.185 & 10.760 & 1.128 & 0.654 \\
      & CNN   & 38.834 & 2.025 & 1.137 & 32.619 & 2.086 & 1.302 & 37.820 & 2.086 & 1.232 \\
      & MLP3  & 17.054 & 1.272 & 0.828 & 20.039 & 1.369 & 0.890 & 21.011 & 1.158 & 0.591 \\
      & MLP5  & 23.735 & 1.902 & 1.464 & 19.996 & 1.629 & 1.069 & 21.794 & 2.072 & 1.666 \\
      \hline
      \multirow{4}{*}{EmpDen\_Des}
      & ANN   & 14.429 & 1.281 & 0.797 & 14.275 & 1.172 & 0.676 & 18.692 & 1.521 & 1.016 \\
      & CNN   & 38.834 & 2.025 & 1.137 & 37.002 & 2.066 & 1.265 & 40.471 & 2.107 & 1.232 \\
      & MLP3  & 17.054 & 1.272 & 0.828 & 16.410 & 1.149 & 0.650 & 18.567 & 1.442 & 1.014 \\
      & MLP5  & 23.735 & 1.902 & 1.464 & 25.323 & 1.908 & 1.549 & 15.156 & 1.199 & 0.687 \\
      \hline
      \multirow{4}{*}{EmpDen\_Ori}
      & ANN   & 14.429 & 1.281 & 0.797 & 15.389 & 1.356 & 0.871 & 12.704 & 1.169 & 0.697 \\
      & CNN   & 38.834 & 2.025 & 1.137 & 34.447 & 2.003 & 1.152 & 32.882 & 2.174 & 1.395 \\
      & MLP3  & 17.054 & 1.272 & 0.828 & 19.749 & 1.157 & 0.656 & 21.386 & 1.243 & 0.743 \\
      & MLP5  & 23.735 & 1.902 & 1.464 & 16.407 & 1.216 & 0.743 & 24.646 & 1.863 & 1.222 \\
      \hline
      \multirow{4}{*}{downtown\_downtown}
      & ANN   & 14.429 & 1.281 & 0.797 & 13.967 & 1.187 & 0.703 & 19.679 & 1.576 & 1.070 \\
      & CNN   & 38.834 & 2.025 & 1.137 & 35.080 & 2.060 & 1.286 & 37.923 & 2.044 & 1.140 \\
      & MLP3  & 17.054 & 1.272 & 0.828 & 17.227 & 1.443 & 0.996 & 17.346 & 1.133 & 0.640 \\
      & MLP5  & 23.735 & 1.902 & 1.464 & 17.658 & 1.120 & 0.622 & 21.022 & 1.392 & 0.776 \\
      \hline
    \end{tabular}%
  }
  \label{tab:all_metrics_lambda_updated}
\end{table*}

\vspace{-2.5em}

\begin{table*}[!t]
  \centering
  \small
  \resizebox{\textwidth}{!}{%
    \begin{tabular}{|l|l|ccc|ccc|ccc|}
      \hline
      \textbf{Feature} & \textbf{Model}
        & \multicolumn{3}{c|}{$\lambda=0.6$}
        & \multicolumn{3}{c|}{$\lambda=0.8$}
        & \multicolumn{3}{c|}{$\lambda=1.0$} \\
      \cline{3-11}
      & & MSE & MAE & MAPE & MSE & MAE & MAPE & MSE & MAE & MAPE \\
      \hline
      \multirow{4}{*}{Commuters\_HW}
      & ANN   & 20.019 & 1.534 & 0.991 & 20.355 & 1.630 & 1.119 & 14.064 & 1.165 & 0.665 \\
      & CNN   & 31.853 & 2.284 & 1.495 & 37.317 & 2.209 & 1.386 & 31.972 & 2.100 & 1.297 \\
      & MLP3  & 22.094 & 1.200 & 0.627 & 16.007 & 1.404 & 1.017 & 15.090 & 1.223 & 0.793 \\
      & MLP5  & 20.449 & 1.285 & 0.667 & 19.442 & 1.233 & 0.627 & 23.160 & 1.394 & 0.805 \\
      \hline
      \multirow{4}{*}{Commuters\_WH}
      & ANN   & 13.926 & 1.306 & 0.783 & 15.248 & 1.292 & 0.795 & 16.856 & 1.381 & 0.857 \\
      & CNN   & 37.593 & 2.151 & 1.328 & 36.625 & 2.088 & 1.299 & 29.327 & 2.064 & 1.285 \\
      & MLP3  & 17.968 & 1.466 & 1.045 & 17.434 & 1.286 & 0.843 & 24.999 & 1.336 & 0.807 \\
      & MLP5  & 20.896 & 1.552 & 1.035 & 17.556 & 1.558 & 1.160 & 17.909 & 1.360 & 0.938 \\
      \hline
      \multirow{4}{*}{EmpDen\_Des}
      & ANN   & 15.558 & 1.219 & 0.725 & 16.067 & 1.257 & 0.776 & 16.525 & 1.410 & 0.923 \\
      & CNN   & 34.403 & 2.131 & 1.336 & 34.358 & 2.107 & 1.314 & 35.578 & 2.162 & 1.339 \\
      & MLP3  & 17.868 & 1.246 & 0.696 & 18.203 & 1.409 & 1.012 & 20.011 & 1.435 & 0.971 \\
      & MLP5  & 25.958 & 1.622 & 0.944 & 23.791 & 1.318 & 0.725 & 22.170 & 1.835 & 1.359 \\
      \hline
      \multirow{4}{*}{EmpDen\_Ori}
      & ANN   & 13.472 & 1.222 & 0.748 & 15.931 & 1.436 & 0.881 & 15.718 & 1.375 & 0.860 \\
      & CNN   & 38.022 & 2.036 & 1.203 & 33.513 & 2.027 & 1.221 & 31.936 & 2.087 & 1.294 \\
      & MLP3  & 15.465 & 1.242 & 0.788 & 18.870 & 1.272 & 0.787 & 21.473 & 1.250 & 0.761 \\
      & MLP5  & 21.074 & 1.436 & 0.806 & 16.441 & 1.469 & 1.093 & 18.098 & 1.654 & 1.332 \\
      \hline
      \multirow{4}{*}{downtown\_downtown}
      & ANN   & 15.519 & 1.310 & 0.812 & 13.943 & 1.240 & 0.769 & 15.199 & 1.355 & 0.842 \\
      & CNN   & 35.785 & 2.086 & 1.241 & 34.547 & 2.179 & 1.383 & 35.026 & 2.100 & 1.280 \\
      & MLP3  & 18.512 & 1.514 & 1.172 & 21.244 & 1.402 & 0.992 & 16.539 & 1.273 & 0.824 \\
      & MLP5  & 35.193 & 1.696 & 1.098 & 28.533 & 2.089 & 1.512 & 23.281 & 1.584 & 1.003 \\
      \hline
    \end{tabular}%
  }
  \caption{MSE / MAE / MAPE for single‐feature experiments across penalty weights.}
  \label{tab:all_metrics_lambda_rest}
\end{table*}

\begin{table*}[!t]
  \centering
  \small
  \resizebox{\textwidth}{!}{%
    \begin{tabular}{|l|l|ccc|ccc|ccc|}
      \hline
      \textbf{Feature} & \textbf{Model}
        & \multicolumn{3}{c|}{$\lambda=0.0$}
        & \multicolumn{3}{c|}{$\lambda=0.2$}
        & \multicolumn{3}{c|}{$\lambda=0.4$} \\
      \cline{3-11}
      & & MSE & MAE & MAPE & MSE & MAE & MAPE & MSE & MAE & MAPE \\
      \hline
      \multirow{4}{*}{$x_1$}
      & ANN  & 0.32057 & 0.45938 & 10.42222 & 0.31106 & 0.45526 & 9.81507  & 0.32053 & 0.45846 & 10.19174 \\
      & CNN  & 0.24235 & 0.39540 & 11.09392 & 0.22121 & 0.37450 & 9.89390  & 0.27737 & 0.41789 & 7.90425  \\
      & MLP3 & 0.26765 & 0.41518 & 9.29944  & 0.26137 & 0.41122 & 10.13378 & 0.23121 & 0.38593 & 6.78722  \\
      & MLP5 & 0.28392 & 0.42149 & 10.75109 & 0.24167 & 0.38163 & 5.07591  & 0.28116 & 0.41926 & 6.31284  \\
      \hline
      \multirow{4}{*}{$x_2$}
      & ANN  & 0.32057 & 0.45938 & 10.42222 & 0.31944 & 0.46012 & 10.43932 & 0.30876 & 0.45403 & 9.88797  \\
      & CNN  & 0.24235 & 0.39540 & 11.09392 & 0.21381 & 0.36665 & 10.32216 & 0.25869 & 0.40256 & 12.01766 \\
      & MLP3 & 0.26765 & 0.41518 & 9.29944  & 0.30772 & 0.44933 & 13.86756 & 0.21875 & 0.37583 & 5.48945  \\
      & MLP5 & 0.28392 & 0.42149 & 10.75109 & 0.25774 & 0.39673 & 5.36131  & 0.26289 & 0.40138 & 4.15501  \\
      \hline
      \multirow{4}{*}{$x_3$}
      & ANN  & 0.32057 & 0.45938 & 10.42222 & 0.31981 & 0.45889 & 9.35833  & 0.31966 & 0.45869 & 9.87373  \\
      & CNN  & 0.24235 & 0.39540 & 11.09392 & 0.25022 & 0.39685 & 9.32460  & 0.24844 & 0.39437 & 13.31762 \\
      & MLP3 & 0.26765 & 0.41518 & 9.29944  & 0.23141 & 0.38938 & 9.29916  & 0.30535 & 0.44834 & 10.19081 \\
      & MLP5 & 0.28392 & 0.42149 & 10.75109 & 0.26257 & 0.40250 & 5.58150  & 0.27335 & 0.41303 & 4.69296  \\
      \hline
    \end{tabular}%
  }
  \label{tab:all_metrics_lambda}
\end{table*}

\vspace{-20.0em}
\begin{table*}[!t]
  \centering
  \small
  \resizebox{\textwidth}{!}{%
    \begin{tabular}{|l|l|ccc|ccc|ccc|}
      \hline
      \textbf{Feature} & \textbf{Model}
        & \multicolumn{3}{c|}{$\lambda=0.6$}
        & \multicolumn{3}{c|}{$\lambda=0.8$}
        & \multicolumn{3}{c|}{$\lambda=1.0$} \\
      \cline{3-11}
      & & MSE & MAE & MAPE & MSE & MAE & MAPE & MSE & MAE & MAPE \\
      \hline
      \multirow{4}{*}{$x_1$}
      & ANN  & 0.32153 & 0.45941 &  9.68727 & 0.31902 & 0.45882 &  8.96072 & 0.31951 & 0.45970 & 10.34450 \\
      & CNN  & 0.21205 & 0.36631 & 12.13348 & 0.21840 & 0.37339 & 10.82129 & 0.23316 & 0.38815 &  8.16293 \\
      & MLP3 & 0.23425 & 0.39344 & 11.27425 & 0.24127 & 0.39450 &  8.29939 & 0.23260 & 0.38896 &  7.07571 \\
      & MLP5 & 0.24905 & 0.39142 &  7.53028 & 0.26218 & 0.40128 &  5.05487 & 0.26920 & 0.40644 &  4.36761 \\
      \hline
      \multirow{4}{*}{$x_2$}
      & ANN  & 0.31628 & 0.45783 & 10.25313 & 0.30870 & 0.45418 & 10.32011 & 0.31677 & 0.45713 &  9.75796 \\
      & CNN  & 0.27189 & 0.41577 & 15.53179 & 0.23741 & 0.39043 & 11.80567 & 0.23005 & 0.38578 &  8.48761 \\
      & MLP3 & 0.30517 & 0.44634 & 12.73852 & 0.25495 & 0.40411 &  6.87999 & 0.30210 & 0.44195 & 11.20974 \\
      & MLP5 & 0.25209 & 0.39589 &  5.71391 & 0.26917 & 0.40961 &  3.98534 & 0.24652 & 0.39158 &  6.68130 \\
      \hline
      \multirow{4}{*}{$x_3$}
      & ANN  & 0.31090 & 0.45433 &  9.97990 & 0.33140 & 0.47272 & 10.35078 & 0.30964 & 0.45275 & 10.22740 \\
      & CNN  & 0.20889 & 0.36579 &  7.18191 & 0.24805 & 0.39926 &  9.92230 & 0.32743 & 0.46648 & 10.36940 \\
      & MLP3 & 0.22133 & 0.37895 &  8.74479 & 0.26710 & 0.41341 &  9.94270 & 0.21520 & 0.37171 &  7.07589 \\
      & MLP5 & 0.27858 & 0.41867 &  6.21411 & 0.27228 & 0.41120 &  5.99696 & 0.26161 & 0.40115 &  7.42778 \\
      \hline
    \end{tabular}%
  }
  \caption{MSE / MAE / MAPE for single‐feature experiments across penalty weights.}
  \label{tab:all_metrics_lambda_rest}
\end{table*}

\begin{table*}[!t]
\centering
\caption{Notation and Symbols}
\footnotesize
\label{tab:notation}
\begin{tabular}{|l|p{0.75\textwidth}|}
\hline
\textbf{Notation} & \textbf{Description} \\
\hline
\multicolumn{2}{|c|}{\textbf{General Symbols}} \\
\hline
$N$ & Number of observations/samples in a batch or dataset \\
$i$ & Index for observations, $i \in \{1, 2, \ldots, N\}$ \\
$j$ & Index for features \\
$\mathcal{J}_M$ & Set of monotonic features (features with expected monotonic relationships) \\
\hline
\multicolumn{2}{|c|}{\textbf{Data and Features}} \\
\hline
$X$ & Input feature matrix \\
$X_j$ & Feature $j$ (column vector of all observations for feature $j$) \\
$X_j^{(i)}$ & Value of feature $j$ for observation $i$ \\
$x_j^{(i)}$ & Sorted/reordered value of feature $j$ for observation $i$ \\
$y$ & True target values \\
$y_i$ & True target value for observation $i$ \\
\hline
\multicolumn{2}{|c|}{\textbf{Model Predictions}} \\
\hline
$\hat{f}$ & Neural network predictions (vector) \\
$\hat{f}^{(i)}$ & Neural network prediction for observation $i$ \\
$\hat{y}_i$ & Predicted value for observation $i$ (same as $\hat{f}^{(i)}$) \\
$\sigma$ & Permutation function for sorting predictions \\
$\hat{f}^{(\sigma(i))}$ & Sorted predictions in ascending order \\
\hline
\multicolumn{2}{|c|}{\textbf{Linear Baseline}} \\
\hline
$g(x)^{(i)}$ & Linear reference prediction for observation $i$ \\
$a_j$ & Slope parameter of linear baseline for feature $j$ \\
$b_j$ & Intercept parameter of linear baseline for feature $j$ \\
$\mu_{X_j}$ & Mean value of feature $j$ \\
$\mu_{\hat{f}}$ & Mean value of neural network predictions \\
$\mathrm{Cov}(X_j, \hat{f})$ & Covariance between feature $j$ and predictions \\
$\mathrm{Var}(X_j)$ & Variance of feature $j$ \\
\hline
\multicolumn{2}{|c|}{\textbf{Monotonicity Violation Measures}} \\
\hline
$\Delta \hat{f}^{(i)}$ & Incremental change in predictions: $\hat{f}^{(i+1)} - \hat{f}^{(i)}$ \\
$\Delta g(x)^{(i)}$ & Incremental change in linear reference: $g(x)^{(i+1)} - g(x)^{(i)}$ \\
$v_j^{(i)}$ & Violation function: $\max\{0, \Delta g(x)^{(i)} - \Delta \hat{f}^{(i)}\}$ \\
$P_j$ & Penalty for feature $j$: $\sum_{i=1}^{N-1}(v_j^{(i)})^2$ \\
\hline
\multicolumn{2}{|c|}{\textbf{Loss Functions}} \\
\hline
$\mathcal{L}_{\text{m}}(\hat{f}, X)$ & Monotonicity penalty term: $\frac{1}{N}\sum_{j \in \mathcal{J}_M}P_j$ \\
$\mathcal{L}(\hat{f}, y)$ & Total loss function: $\text{MSE} + \lambda \cdot \mathcal{L}_{\text{m}}(\hat{f}, X)$ \\
$\lambda$ & Penalty weight/regularization parameter \\
$\lambda_{\text{base}}$ & Base penalty weight for adaptive weighting \\
$\lambda_{\text{adaptive}}$ & Adaptive penalty weight \\
\hline
\multicolumn{2}{|c|}{\textbf{Evaluation Metrics}} \\
\hline
$\text{MSE}$ & Mean Squared Error: $\frac{1}{N}\sum_{i=1}^N (\hat y_i - y_i)^2$ \\
$\text{MAE}$ & Mean Absolute Error: $\frac{1}{N}\sum_{i=1}^N |\hat y_i - y_i|$ \\
$\text{MAPE}$ & Mean Absolute Percentage Error: $\frac{100}{N}\sum_{i=1}^N \frac{|\hat y_i - y_i|}{y_i}$ \\
\hline
\multicolumn{2}{|c|}{\textbf{Gradient Terms}} \\
\hline
$\theta$ & Model parameters \\
$\nabla_{\theta}\mathcal{L}_{\text{main}}(\hat{f}, y)$ & Gradient of main loss with respect to model parameters \\
$\nabla_{\theta}\mathcal{L}_{\text{m}}(\hat{f}, X)$ & Gradient of monotonicity penalty with respect to model parameters \\
$\epsilon$ & Small constant to prevent division by zero \\
$\|\cdot\|_2$ & L2 norm \\
\hline
\multicolumn{2}{|c|}{\textbf{Synthetic Dataset}} \\
\hline
$x_1, x_2, x_3, x_4$ & Synthetic features: $x_1\sim\mathcal{U}(0,200),\;x_2\sim\mathcal{U}(0,50),\;x_3\sim\mathcal{U}(0,150),\;x_4\sim\mathcal{U}(0,100)$ \\
$f_j(x_j)$ & Baseline transforms: $f_1(x_1)=0.5x_1,\;f_2(x_2)=1.2\sqrt{x_2},\;f_3(x_3)=2.0\log(1+x_3),\;f_4(x_4)=-0.8x_4$ \\
$B$ & Number of equal-width bins for piecewise bump injection \\
$b$ & Bin index: $b=\big\lfloor B\,(x_j-\min x_j)/(\max x_j-\min x_j)\big\rfloor$ \\
$\delta_{j,b}$ & Random bump offset in bin $b$: $\delta_{j,b}\sim\mathcal{N}(0,\sigma_j^2)$ \\
$\Delta_j(x_j)$ & Piecewise constant bump function: $\Delta_j(x_j)=\delta_{j,b}$ \\
$\varepsilon_i$ & Gaussian noise term: $\varepsilon_i\sim\mathcal{N}(0,10^2)$ \\
$y_i$ & Target value: $y_i=\displaystyle\sum_{j=1}^4\bigl(f_j(x_{i,j})+\Delta_j(x_{i,j})\bigr)\;+\;\varepsilon_i$ \\
\hline

\end{tabular}
\end{table*}


\end{document}